\documentclass{bjmc}   

\usepackage{natbib}
\usepackage{graphicx}
\usepackage{xcolor}
\usepackage{amsfonts}

\begin{document}

%
\title{DINO Pre-training for Vision-based End-to-end Autonomous Driving
}
\titlerunning{DINO pre-training for end-to-end autonomous driving}  
%
%
\author{Shubham Juneja\inst{1}, Povilas Daniu\v{s}is\inst{2}, Virginijus Marcinkevi\v{c}ius\inst{1} 
}
\institute{Vilnius University Institute of Data Science and Digital Technologies, Akademijos str. 4, Vilnius, LT-08412, Lithuania 
\and
Research Institute of Natural and Technological Sciences, Vytautas Magnus University, 53361 Kaunas, Lithuania
}
%
\emails{shubham.juneja@mif.stud.vu.lt,
povilas.daniusis@vdu.lt,
virginijus.marcinkevicius@mif.vu.lt}
\orcids{ORCID 0000-0002-7906-5688, 0000-0001-5977-827X, 0000-0002-2281-4035 }
%
%

\maketitle      

\begin{abstract} 

In this article, we focus on the pre-training of visual autonomous driving agents in the context of imitation learning.
Current methods often rely on a classification-based pre-training, which we hypothesise to be holding back from extending capabilities of implicit image understanding.
We propose pre-training the visual encoder of a driving agent using the self-distillation with no labels (DINO) method, which relies on a self-supervised learning paradigm.
Our experiments in CARLA environment in accordance with the Leaderboard benchmark reveal that the proposed pre-training is more efficient than classification-based pre-training, and is on par with the recently proposed pre-training based on visual place recognition (VPRPre).


\end{abstract}
%
%





\section{Introduction}

While autonomous driving of robots and vehicles can be achieved by breaking down the task of driving into individual sub-tasks and assigning a module for each, end-to-end learning takes the holistic approach.
End-to-end learning of driving often relies upon imitation learning, i.e., given a corpus of data (for e.g., visual demonstrations), a task is learned with a machine learning method, e.g., neural network. 
Hence, instead of singularly learning every sub-task or programming it into a module, an entire skill is learned in the form of a policy from data. 
To absorb the precise behaviour from data (or demonstrations) needed to safely drive in the real world, datasets of vast sizes may be required along with stronger methods that learn from what is available.
This is one of the issues that has stalled progress in research on purely vision-based end-to-end models \citep{CILplus}, giving rise to hybrid methods combining end-to-end approaches with modular pipelines.

The major cause of insufficient driving performance in imitation learning-based methods stems from the problem of co-variate shift.
Co-variate shift \citep{DAgger, EndtoendSurvey} is caused by occurrences of situations (or data points) at test time which have not been presented at the time of training, resulting in encountering a shift in data distributions.
Specifically in the context of autonomous driving, this can be a mixture of unseen weather conditions, towns, traffic situations, and so on.
While the various methods that propose aggregating of new and vital data points \citep{DAgger,zhang2017querySAFEDAgger,prakash2020exploringDARB} into the existing data corpus have become an essential practice in imitation learning, there still seems to be a lot of space for the learned methods to adapt better.
Besides improving data quality with data aggregation, there are other important lines of research, one of which focuses on model parameter initialisation or pre-training \citep{zhang2022actionACO, wu2023policyPPgeo, VPRpretrain}.

Efficient pre-training of a learning method (e.g. neural network) potentially implies parameter initialisation, that may be related to the learning task of interest.
This has recently been the highlight of advances in language modelling  \citep{minaee2024LLM}, and very commonly used in vision based learning tasks as well (i.e. models pre-trained on ImageNet \citep{deng2009imagenet} such as ResNet \citep{Resnet}).
Majority of the works in autonomous driving rely on a classification-based pre-training 
\citep{codevilla2019exploring,zhang2021endRoach,chitta2022transfuser,CILplus,jia2023thinktwice} and only a handful of works investigate the impact of various other pre-training methods \citep{zhang2022actionACO, wu2023policyPPgeo, VPRpretrain}.

A recent self-supervised method called self-distillation with no labels (DINO) \citep{dinoV1} has shown an inherent understanding of the semantic information of an image, which implies it's potential usefulness for various computer vision tasks, including autonomous driving.

We hypothesise that pre-training a driving agent's vision encoder with heavy guidance based on labels may be indirectly or directly holding back the agent's driving performance when it comes to generalisation.
Such issues could arise due to the presence of strong image-level supervision of supervised methods potentially reducing the concept of learning to a single task.
This may also hold true in some very recent self-supervised pre-training methods \citep{wu2023policyPPgeo,zhang2022actionACO,VPRpretrain}.

We empirically investigate the above hypothesis by applying DINO pre-training \citep{dinoV1}, and comparing it with the standard supervised classification-based pre-training approach and with a recent method, visual place recognition pre-traing for driving agents (VPRPre)
\citep{VPRpretrain}. The contributions of this paper are as follows:
\begin{enumerate}

    \item We propose and empirically investigate DINO pre-training for imitation learning-based autonomous driving agents.
    \item Following the offline Leaderboard benchmark standard \citep{zhang2021endRoach, MILE} on the CARLA 0.9.11 simulator, we empirically demonstrate that the vision encoder pre-trained with downstream-task-agnostic DINO exhibits improved driving performance compared to vision encoders pre-trained with supervised learning (ImageNet classification). Furthermore, our experimental findings reveal that its performance is comparable to the VPRPre encoder \citep{VPRpretrain}, which was also trained in the same CARLA environment.
    
\end{enumerate}

\section{Related work} 


The core concept of using neural networks for autonomous driving in research was initially demonstrated by the ALVINN method \citep{Pomerleau1988ALVINNAA}, and was revisited during the recent connectionist renaissance with the work of \citep{BojarskiTDFFGJM16}.

PilotNet took advantage of a deeper neural network architecture and the capability of available computing power to train it, establishing higher performance.
The use of better and well-adapted architectures for driving has thereon been a strong line of research (e.g., \citep{codevilla2019exploring, Daniusis2021, chitta2022transfuser,CILplus,yokoyama2024vlfm}).
The most notable architecture that has been frequently used and has shown a remarkable improvement in adapting for driving is conditional imitation learning with ResNet (CILRS) \citep{codevilla2018end,codevilla2019exploring}, where each command is given a different multi-layer perceptron branch.
Another line of research that has contributed to the progress of autonomous driving in order to reduce the effect of co-variate shift, is on how to aggregate data better into the training data corpus.
While the core method of data aggregation (DAgger) \citep{DAgger} has brought improvement by simply aggregating corrective data where an agent misbehaves, other DAgger methods \citep{zhang2017querySAFEDAgger,prakash2020exploringDARB} have explored how can that be conducted more efficiently.

Various other aspects of end-to-end autonomous driving have been challenged in order to find ways to enhance its capabilities. 
A milestone in this research was reached by the method called Roach \citep{zhang2021endRoach} which questions the quality of demonstrations used for training, be it human-driven or rule-based demonstration.
The researchers proposing Roach argue that the current form of demonstration may not be well informed, and hence propose a reinforcement learning agent that drives based on a birds-eye view and generates higher quality demonstrations, resulting in better data quality for a latter agent trained over these demonstrations.
The latter agent drives on frontal camera view just as previously mentioned methods, resulting in improved performance.
Meanwhile, CIL++ \citep{CILplus} proposes enriching the vision in the imitation learning-based agent rather than demonstrations, with a higher field of view provided by two additional cameras.
CIL++ also extends on the original CILRS \citep{codevilla2019exploring} method by the use of transformer \citep{vaswani2017attention} architecture to fuse multiple views.
Transfuser \citep{chitta2022transfuser} is another method that also uses a transformers-inspired architecture, and does so to explore multi-modality by extending image-based vision with LiDAR.
Multi-modality for driving has also been previously explored with different architectures \citep{MultimodelYiXao, PerceiverSelf}.
While most methods use a similar architecture for driving, the following method named think twice before driving \citep{jia2023thinktwice} brings emphasis onto the decoder part of the architecture.
This method modifies the decoder to be able to focus on different areas of the input image given the current context.
It makes several coarse predictions and gradually refines the offset to each prediction.

All previously mentioned methods have a high dependence on vision as a modality, and in their basic forms, they utilise a vision encoder pre-trained on ImageNet \citep{deng2009imagenet} in a supervised way.
Rather than directly learning weights from the task of driving, utilising a pre-trained vision encoder provides potentially more advanced starting point for learning.
However, only a handful of works in the area of end-to-end autonomous driving have experimented with other forms of pre-training than the standard ImageNet classification-based pre-training, hence this line of work remains underexplored.
As an example, a recent method pre-trains the vision encoder on the task of visual place recognition at first and then incorporates it into a full-fledged architecture for the task of driving \citep{VPRpretrain}.

While VPRPre \citep{VPRpretrain} corresponds to supervised learning, some recent methods take advantage of the self-supervised learning \citep{zhang2022actionACO, wu2023policyPPgeo}.
Policy pre-training via geometric modelling (PP-Geo) \citep{wu2023policyPPgeo} learns geometric information such as pose, depth and future ego-motion in a self-supervised manner as such information can be made available through a simulator during data collection, while labelled information can have high costs.
Another self-supervised method called action conditioned contrastive pre-training (ACO) \citep{zhang2022actionACO} explores pre-training on contrastive representation learning over YouTube videos as a pre-training method.
Being based on self-supervised learning, both PP-Geo and ACO are heavily guided by labels which may keep these methods from learning a wider set of features and focus only on the task at hand during pre-training.


Supervised approaches often require labelled data which can be expensive to scale.
In contrast, self-supervised learning alleviates this by learning from alternately available meta-data rather than expensive labels, proving to be highly sample efficient.

We base our research on the recently proposed work, DINO \citep{dinoV1}, which trains over ImageNet using a self-supervised approach without the use of manual annotations.
This is conducted with a multi-crop training approach applied onto a contrastive loss, in the presence of a momentum encoder \citep{he2020momentum}.
The authors explore and confirm the results on convolutional networks as well as on transformer networks.
This method has been able to show that learning features in the proposed non-label guided self-supervised way can inherently enable scene layout and object boundaries understanding without any explicit labels for the same.
DINO enables possibilities of exploring in similar directions as PP-Geo, ACO and VPRPre methods, and additionally investigating if supervised ImageNet pre-training may be a practice that can be considered outdated.


\section{Method}

End-to-end autonomous driving based on imitation learning is achieved by training over multiple sets of demonstrations.
As the standard procedure followed by most imitation based techniques, training over a fixed set of demonstrations doesn't suffice, hence after the first training round another set of demonstrations are aggregated \citep{DAgger}.
This is done up-to 5 times and the results of the final iteration is reported.
We aim to pre-train the vision encoder using a non-label guided self-supervised (DINO) method over a general task and then to train a visual end-to-end autonomous driving model relying on the aforementioned DAgger approach.
We reveal the details of our method in the following subsections.


\subsection{DINO Pre-training}

Self-supervised training uses unlabelled data and the artificial supervision signal, provided by the learning algorithm.
We further utilise the DINO \citep{dinoV1} method which performs self-supervised learning over the ImageNet dataset consisting of $\approx 1$ million images, resulting in efficient image representations, that are useful for a variety of down-stream tasks.

DINO uses two networks, a student and a teacher architecture, with same number of parameters that use distillation during training. 
The student network $g_{\theta_s}$ with parameters $\theta_s$ is trained to match the output of a teacher network $g_{\theta_t}$ with parameters $\theta_{t}$.
For an input $x$, both student and teacher networks infer $K$-dimensional probability distributions, $P_s$ and $P_t$ respectively.
Following that, the probabilities are calculated from the distributions using a softmax function with a modification where the sharpness of the distributions are controlled with a temperature parameter.
For the case of the student network, the modified softmax equation can be seen in equation \ref{eq:SoftmaxWithTemp}, where to calculate the probability $P_s$ temperature parameter $\tau_s$ is used to control sharpness.
\begin{equation}
\label{eq:SoftmaxWithTemp}
P_s(x)^{(i)} = \frac{\exp\left(g_\theta(x)^{(i)} / \tau_s\right)}{\sum_{k=1}^K \exp\left(g_\theta(x)^{(k)} / \tau_s\right)}
\end{equation}

A similar equation is used for calculating $P_t$ with temperature parameter $\tau_t$.
The temperature control parameters are conditioned $\tau_s>0$, $\tau_t>0$.

The teacher network is co-trained along with the student network but is frozen during an epoch.
Instead, the exponential moving average of the weights is copied from the student network to the teacher network, using the momentum encoder technique \citep{he2020momentum}.
The update rule used is
\begin{equation}
\label{eq:updateRule}
    \theta_t \leftarrow \lambda \theta_t + (1 - \lambda) \theta_s,
\end{equation}
where $\lambda$ follows a cosine schedule from $0.996$ to $1$ during training.
With the use of a fixed teacher network within an epoch, the learning takes place by minimising cross-entropy w.r.t. the student network parameters $\theta_s$, as in the following equation,
\begin{equation}
\label{eq:crossEntropy}
    \min_{\theta_s} H(P_t(x), P_s(x)),
\end{equation}
where $H(a, b) = -a \log b$.

To leverage the self-supervision, DINO uses multi-crop training \citep{caron2020unsupervised}. 
At first, a set of multiple views or crops $V$ of an image are formed, in two settings. 
First setting creates two views called global views $x^g_1$ and $x^g_2$, which are crops at resolution of $224 \times 224$ that cover more than $50\%$ of the image.
The second setting creates several views called local views which are of resolution $96\times 96$ that cover less than $50\%$ of the image.
Once the views are created, the global views are passed through the teacher network, and the local views are passed through the student network.
Then modified version of the loss in eq. \ref{eq:crossEntropy} is used to adapt to a self-supervised setting in the following way:
\begin{equation}
\label{eq:selfsupervisedCrossEntropy}
\min_{\theta_s} \sum_{x \in \{x_1^{g_1}, x_2^{g_2}\}} \sum_{x' \in V, x' \neq x} H(P_t(x), P_s(x'))
\end{equation}

The neural networks $g_{\theta}$ are composed of a backbone $f$ and projection head $h$. DINO features are represented by the outputs of backbone of student network.



\subsection{Driving}


For the driving agent we follow the framework set in our previous work \citep{VPRpretrain}.
The decoder of the architecture is based upon CILRS \citep{codevilla2019exploring} as commonly improved in many other works \citep{zhang2021endRoach, VPRpretrain}, where high-level command is given by the navigation system activates the corresponding branch of the decoder. 
This high-level command may be one of several discrete instructions, for example, follow lane, turn right, etc.
To collect the initial demonstrations for the base training data, we use Roach \citep{zhang2021endRoach}, which enables automated data collection with a reinforcement learning agent (Roach agent), that drives from bird's eye perspective.
While the agent drives, it collects images from the front camera of the car along with the low-level driving commands executed.
Once we have the initial dataset of demonstrations, we train our agent with a pre-trained encoder integrated into it.
The trained agent is then let to drive in the simulated environment with training settings, and while this trained agent drives it is supervised by the Roach agent.
Hence, at instances where the trained agent makes mistakes i.e. disagrees with the supervising Roach agent, it is corrected by the Roach agent and these instances of the demonstrations are saved for the next iteration of DAgger.
This is followed by training over the aggregated set of corrected demonstrations and the initial dataset together.
This process of collecting aggregated data and re-training is performed for a total of 5 times, as per the benchmark standard \citep{zhang2021endRoach,MILE}.

Similar to recent works \citep{codevilla2019exploring,zhang2021endRoach,VPRpretrain}, our agent's architecture consists of a pre-trained vision encoder that encodes the front-view RGB image, along with a measurements encoder that encodes the current speed of the vehicle and the high-level command from the planner that is one hot encoded.
Both of the encodings are then concatenated and downsized using a join module, formed by fully connected layers.
The output of the join module is ran through the action branches, where each branch is a module of fully connected layers and is responsible for each high-level command.
Based on the high-level command, the corresponding branch's prediction is chosen.
That prediction represents the low-level driving command.
For training, non-corresponding branches are zeroed out.

To mathematically represent our agent, let $\mathit{X} \in \mathbb{R}^{224 \times 224 \times 3}$ be the front-view image from the vehicle.
Thereon, $\mathit{f_{E}}$ being the image encoder with parameters $\theta$ pre-trained using the DINO method, 
$\mathit{u}$ being the vector holding measurements (current speed and high-level command), $\mathit{f_{M}}$ denoting the measurements encoder network with parameters $\xi$, $\mathit{f_{J}}$ denoting the join module with parameters $\phi$ that concatenates the image and measurements encodings and passes through for downsizing, $\mathit{f_{A}}$ being the action branches module with parameters $\psi$ which calculate a low-level command for each high-level command, gives the representation of the network as
\begin{equation}
\label{eq:notaion_one}
    \mathit{f_{A}(f_{J}(f_{E}(X|\theta), f_{M}(u|\xi)|\phi)|\psi)}.
\end{equation}
As the network gives out low-level commands for all possible high-level commands, to select as per the high-level command of interest, let $\mathit{c_{i}}$ be the one-hot encoded command which is indexed with $\mathit{i}$ that zeroes out the non-command branches, we reformulate the network in statement \ref{eq:notaion_one} into an equation as
\begin{equation}
\label{eq:actionEquation}
    \mathit{
    \widehat{\mathbf{a}}(X, u|\theta,\xi,\phi,\psi) := \sum_{i=0}^{n} c_{i}  b_{i}(X, u|\theta,\xi,\phi,\psi),
    }.
\end{equation}
where $b_{i}$ corresponds to the output of $i$-th branch.

For simple comparability with a baseline method, we adapt to the standard loss function used for the task of end-to-end driving \citep{codevilla2019exploring, zhang2021endRoach}, which is the sum of action loss $\mathcal{L}_{A}$ and a speed prediction regularisation $\mathcal{L}_{S}$,
\begin{equation}
\label{eq:PolicyLossFunctions}
\mathcal{L}_{Agent}(\theta,\xi,\phi,\psi) = \mathcal{L}_{A}^{}(\theta,\xi,\phi,\psi) + \lambda_{S} \cdot   \mathcal{L}_{S}^{}.
\end{equation}
Action loss $\mathcal{L}_{A}$ is given by, 
\begin{equation}
\label{eq:ActionLoss}
\mathcal{L}_{A} = \left \|  \mathit{\widehat{\mathbf{a}}(X,u|\theta,\xi,\phi,\psi)} - \mathbf{a} \right \|_{1},
\end{equation}
which calculates the L1 loss between the expert action ${\mathbf{\hat{a}}}$ and learned method's predicted action ${\mathbf{a}}$.
The speed prediction regularisation $\mathcal{L}_{S}$ is given by,
\begin{equation}
\label{eq:SpeedLoss}
\mathcal{L}_{S} =  \mathit{\left |  \hat{s} - s \right |}.
\end{equation}
that calculates the difference between measured speed $\mathit{\hat{s}}$ and predicted speed $\mathit{s}$, and is regulated by a scalar value $\lambda_{s}$ mentioned in eq. \ref{eq:PolicyLossFunctions}.


\section{Experiments}

\subsection{Implementation details}
\label{sec:ImplementationDetails}
To quantify the impact of using DINO pre-training, by following the framework of our previous work \citep{VPRpretrain} we also implement a baseline method.
The baseline method follows the standard setting as in most works that use a convolutional neural network, i.e. it uses a ResNet50 encoder.
This encoder is pre-trained with supervised learning over the ImageNet dataset.

The DINO pre-trained method and the baseline contain exactly the same number of parameters, only differ in the values (or weights) they hold.
While rest of the network is randomly initialised to be trained from scratch.
We initialise the measurement encoder $\mathit{f_{M}}$ with $2$ fully connected layers and set the output dimension to $128$ at each layer.
The join module $\mathit{f_{J}}$ is initialised with 3 fully connected layers and has the output dimensions set to $512$, $512$ and $256$ respectively.
These layers are followed by the action branches $\mathit{f_{A}}$ which consist of 3 fully connected layers with the output dimensions $256$, $256$ and $2$, respectively.
All modules consisting of fully connected layers use a rectified linear unit activation, except the last layers in action branches.

Both methods are trained on the same initial dataset collected with the Roach agent.
We then collect additional data on every trained model, following the formal DAgger procedure.
We iterate with DAgger for 5 times in total and collect $5$ datasets in addition to the common initial dataset, for each method.
Both methods are trained for $20$ epochs, with initial learning rate of $1e-4$ and later stepped down to $1/10^{th}$ of initial value $15^{th}$ epoch onwards.
The training is carried out out on a single Nvidia RTX $3090$ GPU with $24$GB of memory fitting a batch size of $256$.
For both methods we uniformly train on smaller resolution images than resolutions used in most methods, following settings of our previous work \citep{VPRpretrain} due to lack of compute and time resources.
As the dataset size scales over DAgger iterations, the training time for each iteration scales from $10$ to $25$ hours.
Additionally, it requires at least $30$ hours to evaluate every agent that has been trained, making to total time spent over all experiments over $2$ months.

\subsection{Benchmark settings}
\label{sec:BenchmarkSettings}

We benchmark both the methods on the offline version of the Leaderboard benchmark \citep{zhang2021endRoach, MILE}.
The Leaderboard benchmark operates in the CARLA 0.9.11 simulator \citep{dosovitskiy2017carla}, which simulates city and high-way like environments for autonomous driving scenarios.
The simulator lets the traffic (road traffic and pedestrian density) and weather conditions be controlled which brings a strong set of possible combinations to test agents on.
The Leaderboard benchmark defines settings for training and testing, where the agent is to be trained over data from $4$ different town environments with a fixed set of weather conditions, and then tested in $2$ unseen environments along with unseen weather conditions.
\begin{table}
\caption{\textbf{Weather conditions used for training, evaluation and testing.}}
\label{table:WeatherConditions}
\centering
\setlength{\tabcolsep}{3pt}
\begin{tabular}{|l|l|l|}
\hline
{Training weathers} & {Evaluation weathers} & {Testing weathers} \\ [0.1em]
\hline
Wet noon  & Wet noon & Wet sunset \\ [0.3em]
Clear sunset  & Clear sunset & Soft rain sunset \\ [0.3em]
Clear noon  &  &  \\ [0.3em]
Hard rain noon  &  &  \\ [0.3em]

\hline
\end{tabular}
\end{table}

\begin{table}
\caption{\textbf{Towns used for training, evaluation and testing.}}
\label{table:Town Distribution}
\centering
\setlength{\tabcolsep}{3pt}
\begin{tabular}{|l|l|l|}
\hline
{Training towns} & {Evaluation towns} & {Testing towns} \\ [0.1em]
\hline
Town 1  & Town 1 & Town 2 \\ [0.3em]
Town 3  & Town 3 & Town 5 \\ [0.3em]

Town 4 - train & Town 4 - train & Town - 4 test \\ [0.3em]
routes & routes & routes \\ [0.3em]
Town 6  & Town 6 & \\ [0.3em]

\hline
\end{tabular}
\end{table}

We state the settings in Tables \ref{table:WeatherConditions} and \ref{table:Town Distribution}.
For the evaluation task
Following the benchmark standard, the agent is run from a given starting point to a given ending point, in a combination of settings of town and weather.
We run the benchmark with the busy traffic setting, as done in other recent works \citep{zhang2021endRoach,MILE,VPRpretrain}.


\subsection{Metrics}
\label{sec:metrics}

To quantify the success rates of the compared methods we assess scores of two metrics, namely route completion and distance completion.
Given a set of routes for every setting, i.e. train, evaluation and testing, route completion is the average percentage of routes completed in that setting.
In the same way, distance completion represents 
the percentage of distance completed to reach the goal, averaged over all routes in a setting.
While route completion measures the agent’s ability to reach the goal, distance completion assesses the extent to which the agent continues to advance, even if it fails to complete the route.

\section{Results}
\label{sec:Results}


For evaluating our proposed method according to the Leaderboard benchmark standard, we progressively produce $6$ trained agents from $5$ iterations of data aggregation and $1$ initial iteration, each for our proposed method and the baseline method, as mentioned in section \ref{sec:ImplementationDetails}.
Each of these trained agents are evaluated under train town-weather conditions (i.e. in familiar settings) and test town-weather conditions (i.e. unfamiliar settings).
Since the simulation sets up the environment assets such as pedestrians and traffic agents at random, we run our evaluations $3$ times with different random seeds.
We then also report and draw conclusions from the average performances over the metrics mentioned in the section \ref{sec:metrics}.
Furthermore, as the recent work VPRPre \citep{VPRpretrain} aligns with the pre-training line of research and was implemented and evaluated under identical settings, we also incorporate its results into our comparison.

\begin{table}
\caption{\textbf{Route completion (\%) of driving agents on training and new (testing) conditions. Highest of all DAgger iterations reported.}}
\label{table:RouteCompletion}
\centering
\setlength{\tabcolsep}{3pt}
\begin{tabular}{|l|l|l|}
\hline
{Pre-training} & {Train town} & {New town} \\ [0.1em]
{Method} & {\& weather} & {\& weather} \\ [0.3em]
\hline
Baseline  & $77.33 \pm 4$ & $53.20 \pm 1$ \\ [0.3em]
VPRPre  & $81.33 \pm 4$ & $60.25 \pm 2$ \\ [0.3em]
DINO (ours)  & $72.67 \pm 3$ & $62.18 \pm 7$ \\ [0.3em]
\hline
\end{tabular}
\end{table}

\begin{table}
\caption{\textbf{ Distance completion (\%) of driving agents on training and new (testing) conditions. Highest of all DAgger iterations reported. }}
\label{table:DistRatio}
\centering
\setlength{\tabcolsep}{3pt}
\begin{tabular}{|l|l|l|}
\hline
{Pre-training} & {Train town} & {New town} \\ [0.1em]
{Method} & {\& weather} & {\& weather} \\ [0.3em]
\hline

Baseline  & $89.36 \pm 2$ & $72.23 \pm 6$ \\ [0.3em]
VPRPre  & $91.97 \pm 3$ & $86.01 \pm 0$ \\ [0.3em]
DINO (ours)  & $86.04 \pm 1$ & $82.67 \pm 6$ \\ [0.3em]
\hline
\end{tabular}
\end{table}


We denote the best of the scores over all DAgger iterations, in Table \ref{table:RouteCompletion} and Table \ref{table:DistRatio}.
Under the routes completion metric in unfamiliar settings (new town \& weather) in Table \ref{table:RouteCompletion}, DINO pre-training tends to perform better than baseline pre-training and VPRPre by $\approx 9\%$ and $\approx 2\%$ on average, respectively.
Whereas under familiar settings (train town \& weather), while baseline pre-training overtakes the performance of DINO pre-training, it shows signs of an over-fit as the baseline method fails to show generalisation in unfamiliar conditions.
This conjecture is also supported by VPRPre's scores.
A similar trend can be seen in Table \ref{table:DistRatio} where the distance completion metric is compared.
DINO pre-trained method is able to complete higher distance than the baseline method, and comes close to VPRPre's completed distance in unfamiliar settings.
Hence with empirically calculated results in Table \ref{table:RouteCompletion} and Table \ref{table:DistRatio}, our hypothesis aligns with the outcomes of the performed experiments.

\begin{figure*} 
\centering
{\includegraphics[width=0.49\columnwidth]{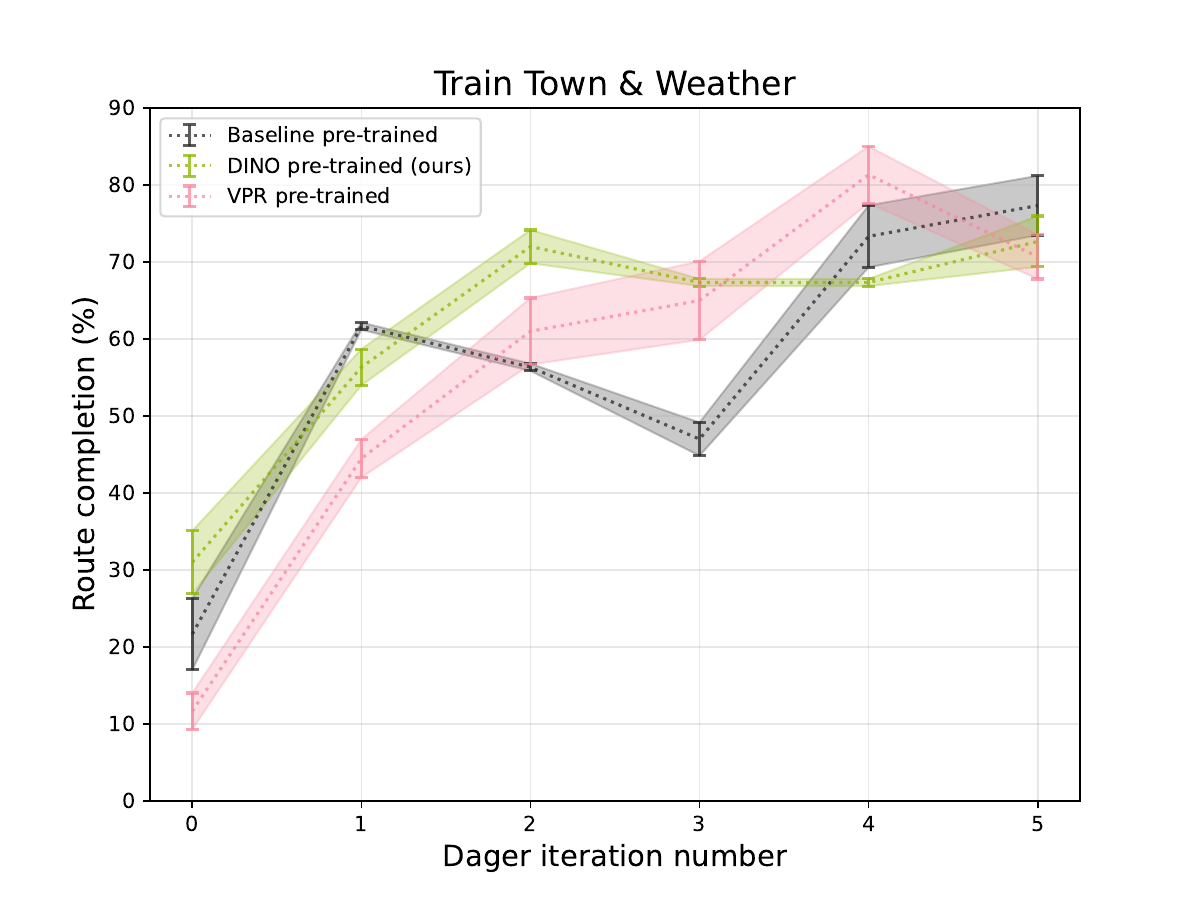}}
\hfil
{\includegraphics[width=0.49\columnwidth]{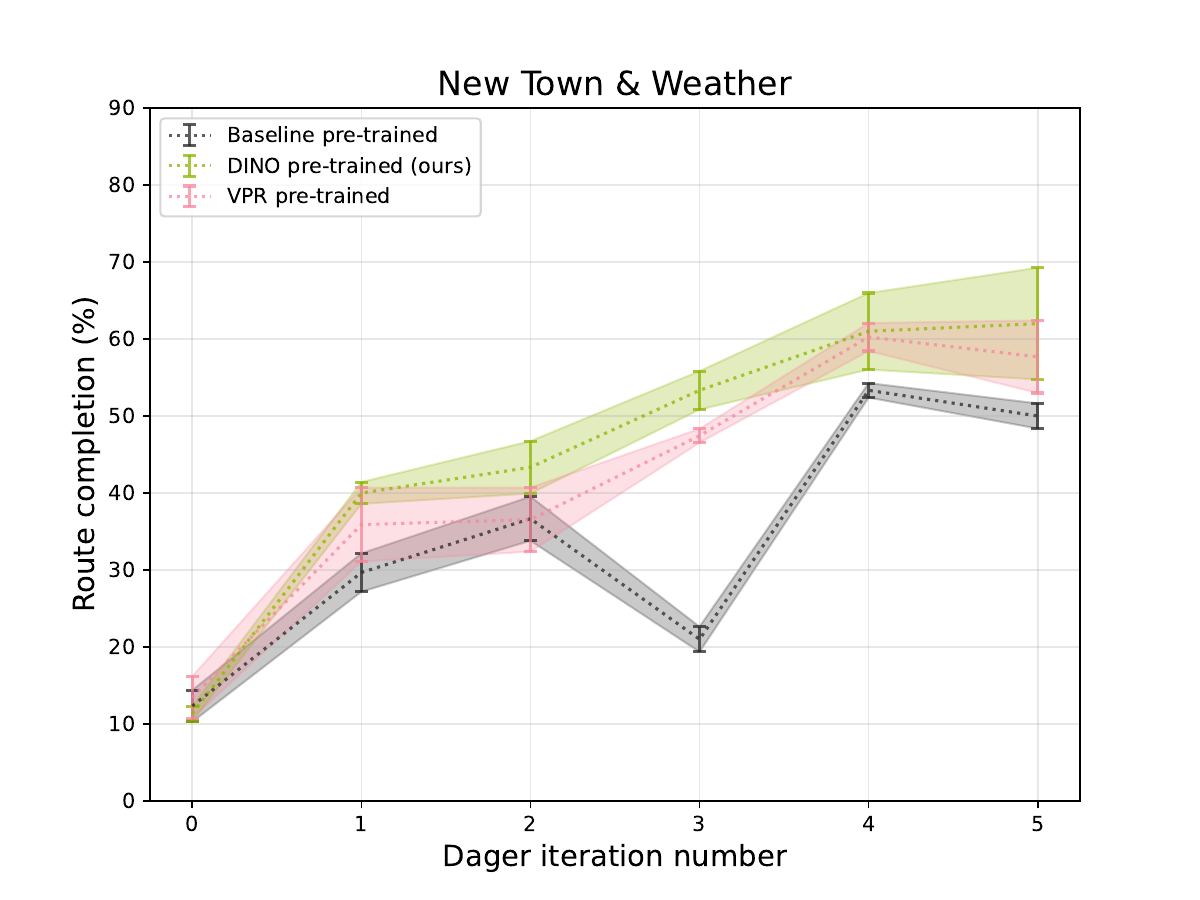}}
\caption{Mean route completion (\%) of evaluating agents over three seeds on the offline Leaderboard benchmark on training conditions (left) and testing conditions (right).}
\label{fig:figure1}
\end{figure*}

\begin{figure*} 
\centering
{\includegraphics[width=0.49\columnwidth]{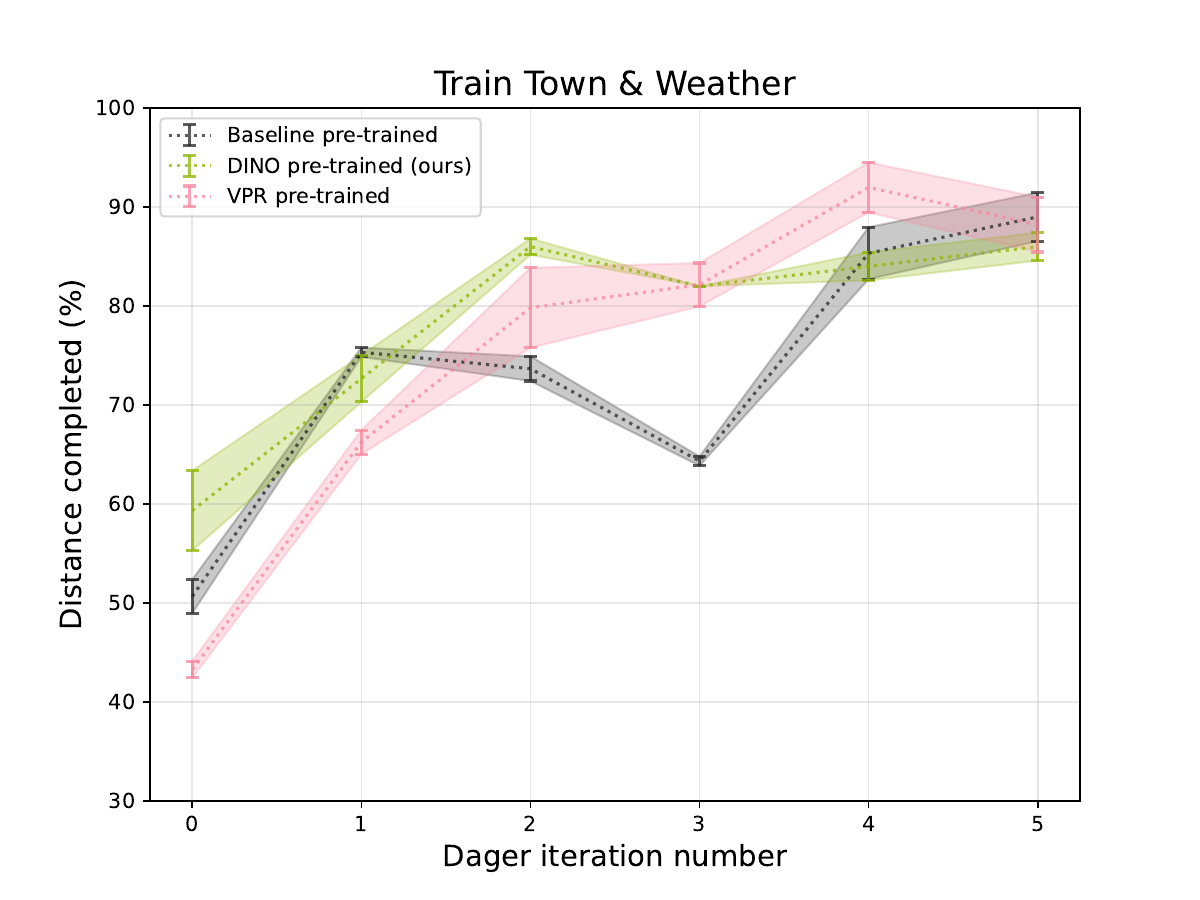}}
\hfil
{\includegraphics[width=0.49\columnwidth]{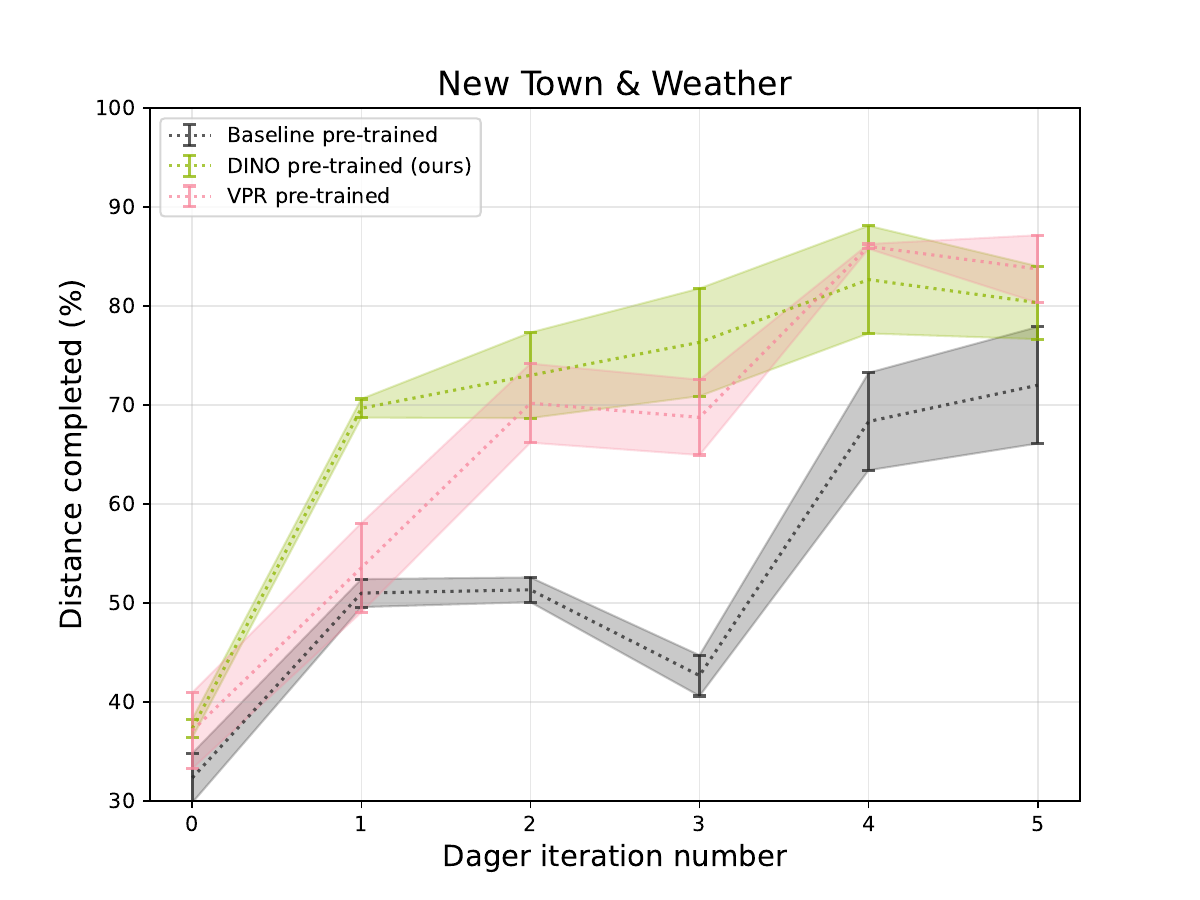}}
\caption{Mean distance completion (\%) of evaluating agents over three seeds on the offline Leaderboard benchmark on training conditions (left) and testing conditions (right).}
\label{fig:figure2}
\end{figure*}

We also compare the scores over both metrics at every iteration of data aggregation, at every random seed and we calculate the means of the random seeds.
This can be seen in Figures \ref{fig:figure1} and \ref{fig:figure2}.
In comparison to the baseline, DINO pre-trained method not only shows better generalisation it shows reduced over-fit and faster convergence.

We further conjecture that the features learned in the encoder while pre-training in a non-label guided self-supervised way are much richer than the features learned while training in a supervised way followed by the baseline method.
The classical way of pre-training over ImageNet dataset with a classification loss may enable a kick-start in learning the task of interest, but due to its sole focus on a single concept of image understanding it does not converge as fast and successfully as method using DINO pre-training, as it can be seen in Figures \ref{fig:figure1} and \ref{fig:figure2}.
VPRPre's results show the ability of going further than our method's covered distances, yet it fails to complete as many routes.
DINO pre-training is based on use of purely general set of training data relying on a much wider distribution, meanwhile VPRPre's pre-training involves exposure to training data captured in the CARLA simulator.
Such exposure can be advantageous to include into DINO's pre-training and further improve results by increasing domain awareness.

Many of the current methods operate with a pre-trained ResNet vision encoder (which we choose as a baseline) and focus on exploring other parts of problem such as the decoder \citep{jia2023thinktwice}, higher field of view with better multi-view fusion \citep{CILplus}, attention enabled multi-modality \citep{chitta2022transfuser}, and so on.
Such an encoder is heavily guided by only classification labels to incorporate image understanding while training.
Our work illustrates that dropping the reliance on such an encoder can be quite beneficial in terms of generalisation to transfer learning task of autonomous driving.
Additionally this work highlights the need of better pre-training while aligning with other works \citep{zhang2022actionACO,wu2023policyPPgeo,VPRpretrain} in this line of research.



\section{Conclusion}

We propose DINO-based pre-training of the vision encoder used for the task of learning end-to-end autonomous driving.
Our experiments reveal that the suggested pre-training is more efficient in unseen environments than the popular classification-based pre-training.
Moreover, DINO-based pre-training is conducted on an unrelated task and its effectiveness comes close to VPRPre \citep{VPRpretrain}, which relies on additional domain awareness coming from its training data.


\bibliographystyle{dcu}		
\bibliography{bibliography}   

@inproceedings{yokoyama2024vlfm,
  title={VLFM: Vision-Language Frontier Maps for Zero-Shot Semantic Navigation},
  author={Naoki Yokoyama and Sehoon Ha and Dhruv Batra and Jiuguang Wang and Bernadette Bucher},
  booktitle={International Conference on Robotics and Automation (ICRA)},
  year={2024}
}

@inproceedings{codevilla2018end,
  title={End-to-end driving via conditional imitation learning},
  author={Codevilla, Felipe and M{\"u}ller, Matthias and L{\'o}pez, Antonio and Koltun, Vladlen and Dosovitskiy, Alexey},
  booktitle={2018 IEEE International Conference on Robotics and Automation (ICRA)},
  pages={4693--4700},
  year={2018},
  organization={IEEE}
}

@article{Daniusis2021,
author={Daniu{\v{s}}is, Povilas
and Juneja, Shubham
and Valatka, Lukas
and Petkevi{\v{c}}ius, Linas},
title={Topological navigation graph framework},
journal={Autonomous Robots},
year={2021},
month={May},
day={04},
volume={45},
pages={633–646},
issn={1573-7527},
doi={10.1007/s10514-021-09980-x},
url={https://doi.org/10.1007/s10514-021-09980-x}
}

@inproceedings{codevilla2019exploring,
  title={Exploring the limitations of behavior cloning for autonomous driving},
  author={Codevilla, Felipe and Santana, Eder and L{\'o}pez, Antonio M and Gaidon, Adrien},
  booktitle={Proceedings of the IEEE/CVF International Conference on Computer Vision},
  pages={9329--9338},
  year={2019}
}

@inproceedings{dosovitskiy2017carla,
  title={CARLA: An open urban driving simulator},
  author={Dosovitskiy, Alexey and Ros, German and Codevilla, Felipe and Lopez, Antonio and Koltun, Vladlen},
  booktitle={Conference on robot learning},
  pages={1--16},
  year={2017},
  organization={PMLR}
}

@article{DAgger,
	author = {Ross, Stephane and J. Gordon, Geoffrey and Andrew Bagnell, J},
	year = {2010},
	month = {11},
	pages = {627-635},
	title = {A Reduction of Imitation Learning and Structured Prediction to No-Regret Online Learning},
	volume = {15},
	_journal = {Journal of Machine Learning Research - Proceedings Track},
  journal = {J. Mach. Learn. Res.}
}

@inproceedings{Pomerleau1988ALVINNAA,
author = {Pomerleau, Dean A.},
title = {ALVINN: An Autonomous Land Vehicle in a Neural Network},
year = {1988},
publisher = {MIT Press},
address = {Cambridge, MA, USA},
booktitle = {Proceedings of the 1st International Conference on Neural Information Processing Systems},
pages = {305–313},
numpages = {9},
series = {NIPS'88}
}

@article{BojarskiTDFFGJM16,
  author    = {Bojarski, Mariusz and Del Testa, Davide and Dworakowski, Daniel and Firner, Bernhard and Flepp, Beat and Goyal, Prasoon and Jackel, Lawrence D and Monfort, Mathew and Muller, Urs and Zhang, Jiakai and others},
  title     = {End to End Learning for Self-Driving Cars},
  journal   = {CoRR},
  volume    = {abs/1604.07316},
  year      = {2016},
  archivePrefix = {arXiv},
  bibsource = {dblp computer science bibliography, https://dblp.org}
}

@inproceedings{prakash2020exploringDARB,
  title={Exploring data aggregation in policy learning for vision-based urban autonomous driving},
  author={Prakash, Aditya and Behl, Aseem and Ohn-Bar, Eshed and Chitta, Kashyap and Geiger, Andreas},
  booktitle={Proceedings of the IEEE/CVF Conference on Computer Vision and Pattern Recognition},
  pages={11763--11773},
  year={2020}
}

@inproceedings{zhang2017querySAFEDAgger,
    author = {Zhang, Jiakai and Cho, Kyunghyun},
    title = {Query-Efficient Imitation Learning for End-to-End Simulated Driving},
    year = {2017},
    publisher = {AAAI Press},
    booktitle = {Proceedings of the Thirty-First AAAI Conference on Artificial Intelligence},
    pages = {2891–2897},
    numpages = {7},
    location = {San Francisco, California, USA},
    series = {AAAI'17}
}

@article{chitta2022transfuser,
  title={Transfuser: Imitation with transformer-based sensor fusion for autonomous driving},
  author={Chitta, Kashyap and Prakash, Aditya and Jaeger, Bernhard and Yu, Zehao and Renz, Katrin and Geiger, Andreas},
  journal={IEEE Transactions on Pattern Analysis and Machine Intelligence},
  year={2022},
  publisher={IEEE}
}

@article{MILE,
  title={Model-based imitation learning for urban driving},
  author={Hu, Anthony and Corrado, Gianluca and Griffiths, Nicolas and Murez, Zachary and Gurau, Corina and Yeo, Hudson and Kendall, Alex and Cipolla, Roberto and Shotton, Jamie},
  journal={Advances in Neural Information Processing Systems},
  volume={35},
  pages={20703--20716},
  year={2022}
}

@inproceedings{zhang2021endRoach,
  title={End-to-end urban driving by imitating a reinforcement learning coach},
  author={Zhang, Zhejun and Liniger, Alexander and Dai, Dengxin and Yu, Fisher and Van Gool, Luc},
  booktitle={Proceedings of the IEEE/CVF international conference on computer vision},
  pages={15222--15232},
  year={2021}
}

@article{zhang2022actionACO,
  title={Learning to Drive by Watching YouTube videos: Action-Conditioned Contrastive Policy Pretraining},
  author={Zhang, Qihang and Peng, Zhenghao and Zhou, Bolei},
  journal={European Conference on Computer Vision (ECCV)},
  year={2022}
}

@inproceedings{wu2023policyPPgeo,
title={Policy Pre-training for Autonomous Driving via Self-supervised Geometric Modeling},
author={Penghao Wu and Li Chen and Hongyang Li and Xiaosong Jia and Junchi Yan and Yu Qiao},
booktitle={International Conference on Learning Representations},
year={2023}
}

@inproceedings{deng2009imagenet,
  title={Imagenet: A large-scale hierarchical image database},
  author={Deng, Jia and Dong, Wei and Socher, Richard and Li, Li-Jia and Li, Kai and Fei-Fei, Li},
  booktitle={2009 IEEE conference on computer vision and pattern recognition},
  pages={248--255},
  year={2009},
  organization={IEEE}
}

@ARTICLE{EndtoendSurvey,
author={Tampuu, Ardi and Matiisen, Tambet and Semikin, Maksym and Fishman, Dmytro and Muhammad, Naveed},
journal={IEEE Transactions on Neural Networks and Learning Systems}, 
title={A Survey of End-to-End Driving: Architectures and Training Methods}, 
year={2022},
volume={33},
number={4},
pages={1364-1384},
doi={10.1109/TNNLS.2020.3043505}}

@inproceedings{Resnet,
  title={Deep residual learning for image recognition},
  author={He, Kaiming and Zhang, Xiangyu and Ren, Shaoqing and Sun, Jian},
  booktitle={Proceedings of the IEEE conference on computer vision and pattern recognition},
  pages={770--778},
  year={2016}
}

@ARTICLE{VPRpretrain,
  author={Juneja, Shubham and Daniušis, Povilas and Marcinkevičius, Virginijus},
  journal={IEEE Access}, 
  title={Visual Place Recognition Pre-Training for End-to-End Trained Autonomous Driving Agent}, 
  year={2023},
  volume={11},
  number={},
  pages={128421-128428},
  doi={10.1109/ACCESS.2023.3331678}}

@inproceedings{dinoV1,
  title={Emerging Properties in Self-Supervised Vision Transformers},
  author={Caron, Mathilde and Touvron, Hugo and Misra, Ishan and J\'egou, Herv\'e  and Mairal, Julien and Bojanowski, Piotr and Joulin, Armand},
  booktitle={Proceedings of the International Conference on Computer Vision (ICCV)},
  year={2021}
}

@misc{CILplus,
 title={Scaling Vision-based End-to-End Driving with Multi-View Attention Learning},
 author={Yi Xiao and Felipe Codevilla and Diego Porres and Antonio M. Lopez},
 year={2023},
 eprint={2302.03198},
 archivePrefix={arXiv},
 primaryClass={cs.CV}
 }

@article{PerceiverSelf,
    author={Juneja, Shubham and Daniušis, Povilas and Marcinkevičius, Virginijus},
    title = {Combining Multiple Modalities with Perceiver in Imitation-based Urban Driving},
    journal = {ALLSENSORS 2021, The Sixth International Conference on Advances in Sensors, Actuators, Metering and Sensing},
    year = {2021}
}

@article{MultimodelYiXao,
author = {Xiao, Yi and Codevilla, Felipe and Gurram, Akhil and Urfalioglu, Onay and L\'{o}pez, Antonio M.},
title = {Multimodal End-to-End Autonomous Driving},
year = {2022},
issue_date = {Jan. 2022},
publisher = {IEEE Press},
volume = {23},
number = {1},
issn = {1524-9050},
url = {https://doi.org/10.1109/TITS.2020.3013234},
doi = {10.1109/TITS.2020.3013234},
journal = {Trans. Intell. Transport. Sys.},
month = {jan},
pages = {537–547},
numpages = {11}
}

@inproceedings{he2020momentum,
  title={Momentum contrast for unsupervised visual representation learning},
  author={He, Kaiming and Fan, Haoqi and Wu, Yuxin and Xie, Saining and Girshick, Ross},
  booktitle={Proceedings of the IEEE/CVF conference on computer vision and pattern recognition},
  pages={9729--9738},
  year={2020}
}

@article{caron2020unsupervised,
  title={Unsupervised learning of visual features by contrasting cluster assignments},
  author={Caron, Mathilde and Misra, Ishan and Mairal, Julien and Goyal, Priya and Bojanowski, Piotr and Joulin, Armand},
  journal={Advances in neural information processing systems},
  volume={33},
  pages={9912--9924},
  year={2020}
}

@inproceedings{jia2023thinktwice,
  title={Think twice before driving: Towards scalable decoders for end-to-end autonomous driving},
  author={Jia, Xiaosong and Wu, Penghao and Chen, Li and Xie, Jiangwei and He, Conghui and Yan, Junchi and Li, Hongyang},
  booktitle={Proceedings of the IEEE/CVF Conference on Computer Vision and Pattern Recognition},
  pages={21983--21994},
  year={2023}
}

@article{vaswani2017attention,
  title={Attention is all you need},
  author={Vaswani, Ashish and Shazeer, Noam and Parmar, Niki and Uszkoreit, Jakob and Jones, Llion and Gomez, Aidan N and Kaiser, {\L}ukasz and Polosukhin, Illia},
  journal={Advances in neural information processing systems},
  volume={30},
  year={2017}
}

@article{minaee2024LLM,
  title={Large language models: A survey},
  author={Minaee, Shervin and Mikolov, Tomas and Nikzad, Narjes and Chenaghlu, Meysam and Socher, Richard and Amatriain, Xavier and Gao, Jianfeng},
  journal={arXiv preprint arXiv:2402.06196},
  year={2024}
}
\end{document}